%% file: lrec-coling2024-example.tex
\documentclass[10pt, a4paper]{article}
\usepackage{lrec-coling2024} % this is the new style
\usepackage{subfiles}
\usepackage{natbib}
\usepackage{multibib}
\usepackage[nottoc]{tocbibind}
\usepackage{multirow}
\makeatletter
\def\@mb@citenamelist{cite,citep,citet,citealp,citealt,citepalias,citetalias}
\makeatother
\newcites{languageresource}{~}
\usepackage{graphicx}
\usepackage{tabularx}
\usepackage{soul}
\usepackage{caption}
\usepackage{blindtext}
\usepackage[normalem]{ulem}
\useunder{\uline}{\ul}{}

\usepackage{titlesec}

\usepackage{xcolor}
\usepackage{hyperref}
 \definecolor{darkblue}{rgb}{0, 0, 0.5}
  \hypersetup{colorlinks=true, citecolor=darkblue, linkcolor=darkblue, urlcolor=darkblue}

\usepackage{xstring}

\usepackage{color}

\title{\textbf{Extracting Social Determinants of Health from Pediatric Patient Notes Using Large Language Models: Novel Corpus and Methods}}

\name{ \parbox{\textwidth}{\fontsize{12}{0}\selectfont\centering
Yujuan Fu\textsuperscript{1*}\footnotemark[1],
Giridhar Kaushik Ramachandran\textsuperscript{2*}\footnotemark[2] ,
Nicholas J Dobbins\textsuperscript{1}\footnotemark[1] \\
Namu Park\textsuperscript{1}\footnotemark[1], 
Michael Leu\textsuperscript{1}\footnotemark[1],
Abby R. Rosenberg\textsuperscript{3,4,5}\footnotemark[3],
Kevin Lybarger\textsuperscript{2}\footnotemark[2],
Fei Xia\textsuperscript{1,6}\footnotemark[4] \\
Özlem Uzuner\textsuperscript{2}\footnotemark[2],
Meliha Yetisgen\textsuperscript{1}\footnotemark[1]
}}

\address{\textsuperscript{1}Department of Biomedical Informatics \& Medical Education, University of Washington \\
\textsuperscript{2}Department of Information Sciences and Technology, George Mason University \\
\textsuperscript{3}Dana-Farber Cancer Institute,
\textsuperscript{4}Boston Children's Hospital, \\
\textsuperscript{5}Harvard Medical School,
\textsuperscript{6}Department of Linguistics, University of Washington \\
         \textsuperscript{1,4}Seattle, WA, USA, 
         \textsuperscript{2}Fairfax, VA, USA,
         \textsuperscript{3,4,5}Boston, MA, USA
         \\
         \{velvinfu, ndobb, npark95, mgl27, fxia, melihay\}@uw.edu\\
         \{gramacha, klybarge, ouzuner\}@gmu.edu, 
         abbyr\_rosenberg@dfci.harvard.edu\\
                  \textsuperscript{*}Authors contributed equally to this paper.
         }

\abstract{
Social determinants of health (SDoH) play a critical role in shaping health outcomes, particularly in pediatric populations where interventions can have long-term implications. SDoH are frequently studied in the Electronic Health Record (EHR), which provides a rich repository for diverse patient data. In this work, we present a novel annotated corpus, the Pediatric Social History Annotation Corpus (PedSHAC), and evaluate the automatic extraction of detailed SDoH representations using fine-tuned and in-context learning methods with Large Language Models (LLMs). PedSHAC comprises annotated social history sections from 1,260 clinical notes obtained from pediatric patients within the University of Washington (UW) hospital system. Employing an event-based annotation scheme, PedSHAC captures ten distinct health determinants to encompass living and economic stability, prior trauma, education access, substance use history, and mental health with an overall annotator agreement of 81.9 F1. Our proposed fine-tuning LLM-based extractors achieve high performance at 78.4 F1 for event arguments. In-context learning approaches with GPT-4 demonstrate promise for reliable SDoH extraction with limited annotated examples, with extraction performance at 82.3 F1 for event triggers.
\\ \newline\Keywords{Social Determinants of Health, Pediatrics, Information Extraction, Large Language Models}}

\begin{document}

\maketitleabstract

\section{Introduction}

\subfile{sections/2_introduction.tex}

\section{Related Work}

\subfile{sections/3_related_work.tex}

\section{Methods}

\subfile{sections/4_methods.tex}

\section{Results}

\subfile{sections/5_results.tex}

\section{Conclusion}

\subfile{sections/6_conclusion.tex}

\section{Acknowledgements}
This work was supported by the National Institutes of Health (NIH) - National Cancer Institute (Grant Nr. 1R01CA248422-01A1 and 1R21CA258242-01) and National Library of Medicine (Grant Nr. 2R15LM013209-02A1). The content is solely the responsibility of the authors and does not necessarily represent the official views of the NIH. 

We extend our heartfelt gratitude to each annotator for their significant contribution on the dataset annotation:  Brianna L Cowin, Sabrina J Crooks, Karina Lopez, Mehar Maju, Kelsi F Nabity, Jocelyn M Waggoner.

\section{Ethics}

\subfile{sections/7_ethics.tex}

%\section{Bibliographical References}\label{sec:reference}

%\nocite{*}
\bibliographystyle{lrec-coling2024-natbib.bst}
\bibliography{lrec-coling2024-example}

%\section{Appendix}

\end{document}

%% file: sections/2_introduction.tex
Health outcomes and quality of life are affected by the conditions in which people work and live and are referred to as Social Determinants of Health (SDoH) \citep{cdc_2021}. SDoH are particularly important in pediatric populations because health disparities have a long-term impact on future attainment of health, including educational and economic success \citep{thompson2016addressing,dickson2023health}. 
Clinicians have continuously adapted practices by systematically gathering pediatric patients' SDoH during clinical consultations \citep{garg2013addressing, ho2016standard, kazak2015psychosocial}. Previous research has identified screening and intervention for SDoH risks in pediatric patients associated with better health outcomes and highlighted the necessity for a more comprehensive SDoH tool \citep{MORONE201722}. 

% is this better?
However, there are difficulties in documenting SDoH in Electronic Health Records (EHRs) in a tabular format, mainly due to the diversity of SDoH determinants, individual determinants' infrequent occurrence, and inconsistent reporting practice \citep{linfield2023evaluating}. Many pediatric SDoH elements are primarily documented within the clinical narratives from EHRs. Such predominance of unstructured SDoH information in the EHRs impedes the systematic collection and utilization of SDoH information in clinical and research settings, limiting the potential for data-driven inventions to improve individual and public health.

To address these challenges, natural language processing (NLP) information extraction (IE) models are needed to extract semantic representations of SDoH, to enable large-scale and real-time use of this information. IE in the clinical domain and, more broadly, in the general domain has predominantly used fine-tuning-based techniques; recent advancements in instruction-tuned large language models (LLMs) \citep{thirunavukarasu2023large}, trained on large data repositories, are enabling in-context learning approaches. 

Although there is a robust body of IE research exploring SDoH for adult populations, including the development of annotated data sets and data-driven IE models, there is comparatively little IE research investigating the SDoH of pediatric patients. In this work, we present the \textbf{Ped}iatric \textbf{S}ocial \textbf{H}istory \textbf{A}nnotation \textbf{C}orpus (\textbf{PedSHAC}), an annotated corpus of ten distinct SDoH determinants on clinical narratives from pediatric patients from the University of Washington (UW) hospital system. This corpus bridges the gaps in the literature by creating a human-annotated comprehensive and fine-grained corpus of SDoH phenomena for pediatric patients.

To the best of our knowledge, our novel pediatric SDoH corpus, PedSHAC, is the first annotated corpus of pediatric clinical narratives to utilize comprehensive and fine-grained SDoH annotations, including assigning SDoH labels such as \textit{Status} and \textit{Type} that could be incorporated into structured data fields within EHRs to represent patient information better. We believe that this corpus will be a valuable resource in support of understanding the role of SDoH in managing children's health and improving outcomes. Using PedSHAC, we explored various LLM-based IE strategies and demonstrated that detailed SDoH representations can be extracted with high accuracy. The de-identified PedSHAC corpus, annotation guideline, and code are made available through our GitHub\footnote{https://github.com/uw-bionlp/PedSHAC}.

%Our proposed approaches reach human-comparable performance across multiple annotated SDoH determinants.
%Both our best fine-tuning- and in-context-learning-based approaches reach human-comparable performance. %With our high-performing fine-tuned models, we extract the annotated SDoH phenomena with 80.9 F1 (event trigger) and 78.4 F1 (event argument) while our best in-context-learning-based few-shot learning model has 82.3 F1 and 71.6 for SDoH event triggers and arguments. 

%% file: sections/3_related_work.tex
Our contributions include a novel corpus of pediatric clinical narratives with fine-grained annotations (PedSHAC) and comprehensive IE model development for benchmarking. To contextualize both contributions, we describe literature related to both SDoH corpora and IE methods.

\subsection{SDoH Corpora}
The interplay of various social and economic factors on patient health has led to an increased interest in investigating SDoH. To facilitate SDoH exploration, multiple SDoH corpora have been developed. However, their annotation schema might have generally lacked granularity and comprehensiveness, or the patient population might have limited extension into the pediatric domain 

For the adult population, many studies have focused on a limited number of SDoH factors with singular focus such as smoking status \citep{uzuner2008identifying, savova2008mayo}, homelessness \citep{gundlapalli2013using, bejan2018mining}, and substance use \citep{wang2015automated, yetisgen2017automatic, carrell2015using, alzubi2022automated}. %adverse childhood experiences \citep{bejan2018mining, wu2022ontologydriven, wu2022adverse}.
Previous research also addresses SDoH factors in specific contexts, such as sexual health \citep{feller2018towards} and hospital readmission rate \citep{navathe2018hospital}. Our prior SDoH work investigated adult SDoH factors using a fine-grained, event-based annotation scheme encompassing detailed status and type labels for adults \citep{lybarger2021annotating}. 

%SDoH related to the pediatric population was mainly explored through work focused on adverse childhood experiences of the adult patient population \citep{bejan2018mining, wu2022ontologydriven, wu2022adverse}. 
Pediatric SDoH factors such as adverse childhood experiences were researched in the adult patient population \citep{bejan2018mining, wu2022ontologydriven, wu2022adverse}.
The rest of prior SDoH work focused on adult populations doesn't necessarily extend to pediatric-patient-focused corpora, because pediatric populations have unique SDoH factors and there are many factors associated with caregivers that impact the SDoH and health of pediatric patients. For example, education access \citep{dejong2016identifying} and food insecurity \citep{BAER2015601} are especially important to pediatric patients. The clinical notes of pediatric patients may describe employment associated with patient caregivers \citep{kuhlthau2001child, xie2023associations}; at the same time, patient parents' mental health \citep{stallard2004effects} become important as pediatricians continually evaluate whether children may be at risk for child abuse and neglect \citep{farrell2017community}. PedSHAC bridges this gap in the literature with comprehensive fine-grained annotation of SDoH determinants with a focus on pediatric patients.

\subsection{Extraction methods}
SDoH IE is an increasingly explored task, and the modeling approaches range from manually curated rules \citep{patra2021extracting, hatef2019assessing}, traditional/shallow machine learning models \citep{clark2008identifying, wang2015automated}, neural networks \citep{bejan2018mining, gehrmann2018comparing}, to transformer-based LLMs \citep{patra2021extracting, bompelli2021social}. 

Bidirectional Encoder Representations from Transformers (BERT) \citep{devlin-etal-2019-bert} is frequently used in SDoH extraction tasks for text classification \citep{yu2021study, yu2022assessing, han2022classifying} and entity and relation extraction \citep{richie2023extracting, lybarger2022mSpERT}. Sequence-to-sequence approaches that utilize generative LLMs, like Text-to-Text Transfer Transformer (T5) \citep{t5}, have also achieved high performance \citep{romanowski2023extracting}. The most recent generation of LLMs, such as GPT-4 \citep{openai2023gpt4}, are pre-trained on large amounts of data and instruction-tuned \citep{ouyang2022training}, enabling prompt-based learning methods with zero or few in-context examples. Recent work demonstrates the use of GPT-based models in few-shot clinical IE \citep{agrawal-etal-2022-large,yang2023evaluations}.

This work explores pediatric SDoH extraction using multiple transformer-based methods, including fine-tuning through BERT- and T5-based models, and in-context learning using GPT-4. Our experiments showed human-comparable performance through fine-tuning and relatively high performance through in-context learning. Our pipeline is versatile and can be readily adapted to various IE tasks, as a reference for the broader research community.

%% file: sections/4_methods.tex
\subsection{Dataset}

%include details about where the notes were sampled from
This work utilized the clinical notes of pediatric patients from the UW hospital system. The patient cohort consists of a random sample from the general pediatric population to improve generalizability across patient demographics. The clinical notes span a ten-year period (1/1/2012-12/31/2021) with 198k distinct notes from 36k distinct patients. 
Clinical notes are organized into topical sections that are delineated by specific heading formats. Patient SDoH can be described throughout the clinical narrative; however, SDoH are most frequently documented in the social history sections of the clinical notes. 
To focus the annotation on SDoH-dense portions of the clinical notes, we applied a rule-based approach to identify topical section headings and the social history sections, yielding 11k social history sections for 8k distinct patients. The social history section text for a patient can be very similar or identical across notes, so we randomly selected one social history section per patient, resulting in 8k patients, each with a single social history section. Finally, we randomly sampled 1,260 out of 8K social history sections for SDoH annotations. % Clinical notes in PedSHAC are well represented across different age groups, emphasizing early childhood and adolescent cohorts. 
% PedSHAC's patient age distribution is visualized in Figure.\ref{age_hist}.

% \begin{figure}[h]
% \centering
% \captionsetup{justification=raggedright,singlelinecheck=false}
% \includegraphics[width=\linewidth]{figures/age_hist.png}
% \caption{Patient age distribution in the PedSHAC corpus.}
% \label{age_hist}
% \end{figure}

\subsubsection{Annotation scheme}
We created detailed annotation guidelines for ten SDoH (referred to here as \textit{event types}), as listed
in Table \ref{ann_guideline}. 
The three substance events,  \textit{alcohol}, \textit{drug}, and \textit{tobacco}, are annotated and evaluated separately, but their performance is reported together due to their relatively low frequency.

Each event is characterized by a trigger and multiple arguments that describe the event's status, type, and status. The \textit{trigger} is a span with an event-type label. Each \textit{argument} attaches to the corresponding trigger and is assigned a multi-class label, referred to here as a \textit{subtype} label \footnote{ \textit{arguments} and \textit{subtype} labels can be considered as attribute names and attribute values. We chose this naming convention following the previous N2C2 SDoH challenge \cite{lybarger20232022}.}, representing \textit{normalized} SDoH concepts (such as \textit{Status} - \textit{past}, \textit{current}) that are more suitable for downstream clinical applications. Because the most important clinical information is usually stored in a structured format in EHRs, the normalized SDoH concepts as labels can be directly added to other structured information to create a more comprehensive patient representation. Arguments can be categorized into \textit{required} and \textit{optional}. The required arguments define the most important attributes of the event. A trigger can only be annotated if all required arguments can be resolved. 
% add table about the age

\begin{figure}[ht]
\includegraphics[width=\linewidth]{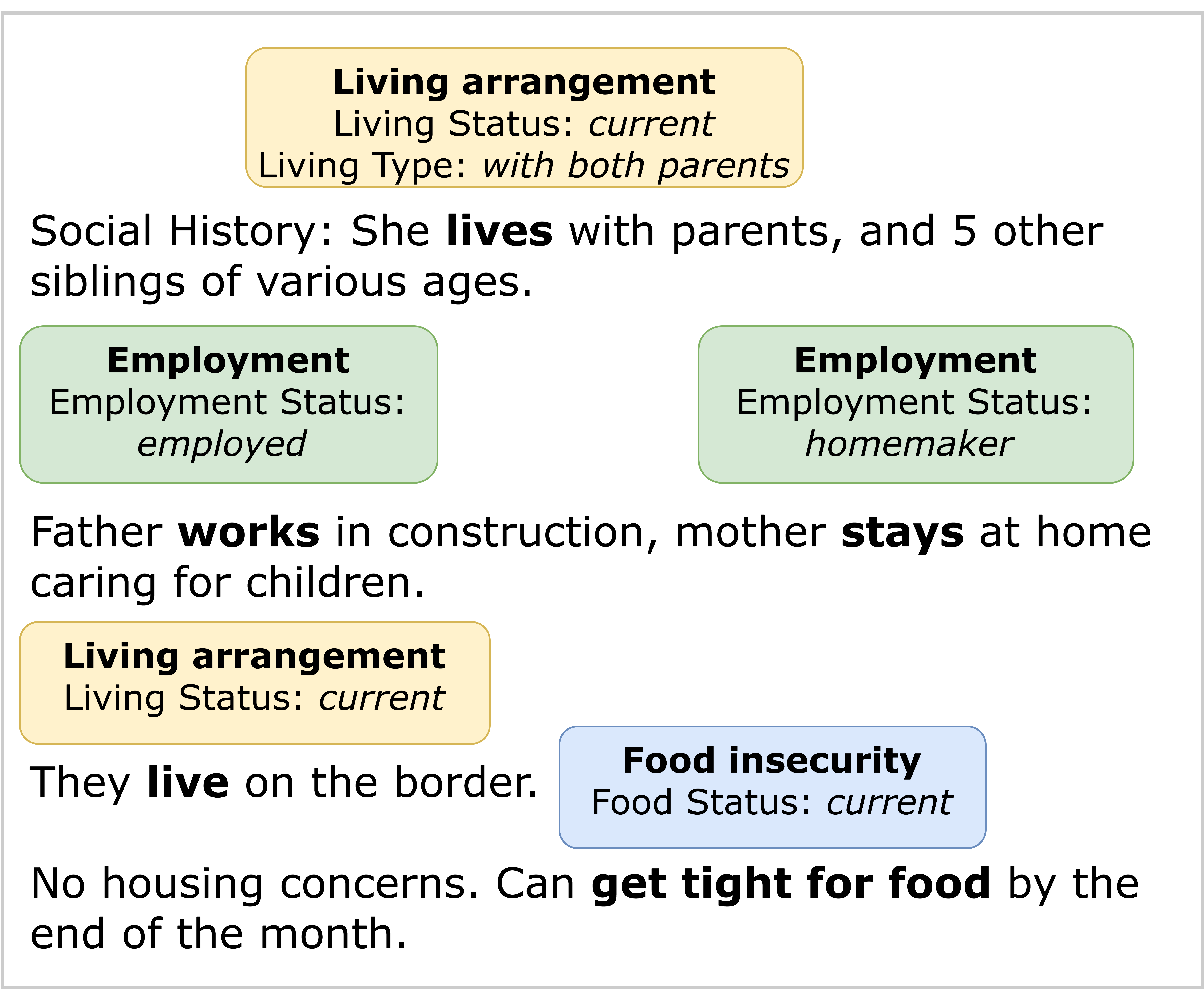}
\captionsetup{justification=raggedright,singlelinecheck=false}
\caption{An Annotation example: the triggers are in boldface. The box above a trigger shows the event type,  arguments and subtype labels.}
\label{ann_example}
\end{figure}

The annotation scheme and event type distribution are specified in Table \ref{ann_guideline}.  SDoH information was annotated using the BRAT rapid annotation tool \citep{stenetorp-etal-2012-brat}. Figure \ref{ann_example} is an example describing the patient's living arrangement and caregivers' employment.

% link to the draw.io: https://drive.google.com/file/d/1LHnD4ReF8qVknT0khVjdGhELcqgkN6-b/view?usp=sharing

% link to the figure:
\begin{figure*}[ht]
    \centering
    \includegraphics[width=\linewidth]
    {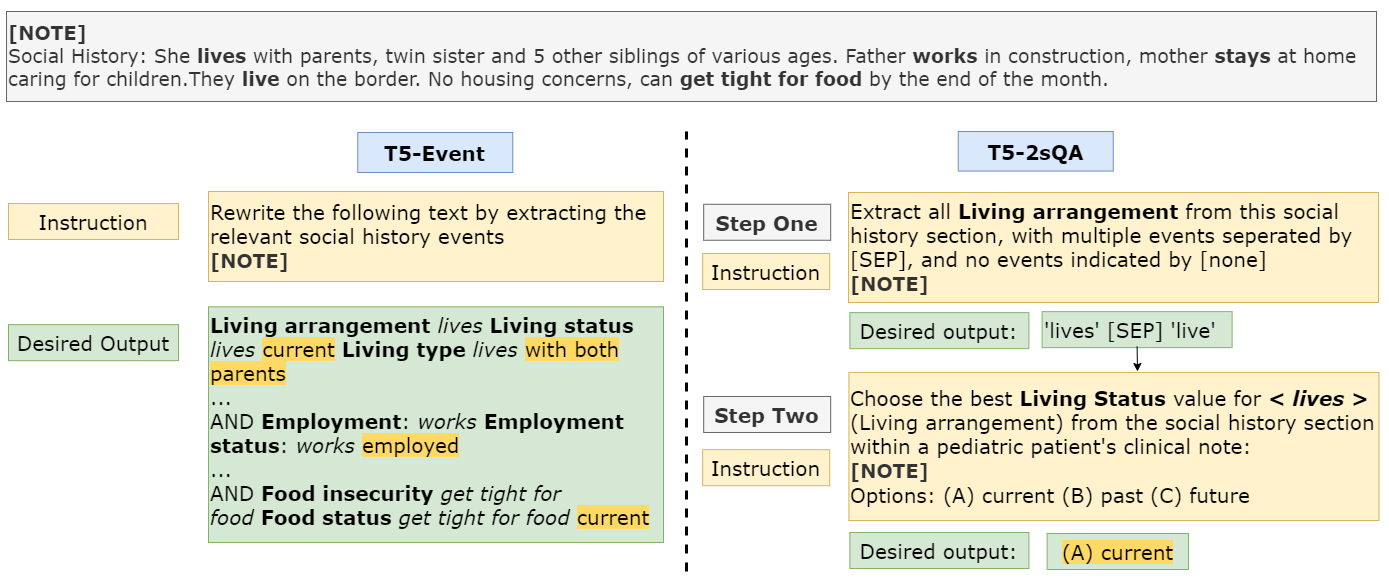}
    \caption{Our one-step (T5-Event) and two-step (T5-2sQA) extraction models. T5-Event extracts all SDoH events, including triggers and arguments, in one query. T5-2sQA extracts triggers and arguments in separate queries, where Step Two includes the predicted triggers from Step One.}
    \captionsetup{justification=raggedright,singlelinecheck=false}
    \label{SDOH_generative_models}
\end{figure*}

\subsubsection{IE evaluation}
\label{evaluation_criteria}
 We follow the previous N2C2 SDoH challenge \citep{lybarger20232022} evaluation criteria. We evaluate the trigger and argument extraction performance for each event. Two triggers are considered \textit{equivalent} if they have the same event type and overlapping spans. The trigger extraction is framed as a named entity recognition task, and the precision, recall, and F1 are calculated. Two arguments are considered \textit{equivalent} if they are attached to equivalent triggers and have the same argument type and subtype labels, and are evaluated using precision, recall, and F1. %For all predicted arguments, we compare them with the gold arguments with equivalent triggers on the subtype label and calculate the precision, recall, and F1, similar to triggers. 

\begin{table*}[ht]
\centering
\small
\input{tables/annotation_guideline.tex}

\caption{Annotation scheme and event statistics for PedSHAC, where * indicates optional arguments. The train, validation, and test sets contain 894, 121, and 245 notes, respectively. The IAA micro-averaged F1 (\%) is calculated on the last round of double annotation, consisting of 90 notes. The IAA F1 micro averages on triggers, arguments, and triggers plus arguments are 85.1, 80.0, and 81.9, respectively.}
\label{ann_guideline}
\end{table*}

\subsubsection{Annotator agreement}
Six medical students at UW annotated SDoH events in our dataset. We first performed two practice rounds to train the annotators and refine the annotation guidelines, with 5 and 10 notes, respectively. After the practice rounds, each note was annotated by two annotators (double annotation), with a third annotator adjudicating disagreements. The Inter-Annotator Agreement (IAA) is evaluated using the criteria in Section \ref{evaluation_criteria}. We doubly annotated 360 notes through 4 rounds (90 notes per round) and then singly annotated the remaining 885 notes. PedSHAC has an IAA micro average of 85.1 F1 across all triggers and 80.0 F1 across all arguments in the last double-annotation round with 90 notes. Low IAAs are from infrequently occurring events such as \textit{Food Insecurity} and \textit{Mental Health}, and the annotation group carefully discussed every disagreement. PedSHAC is split into training, validation, and test sets. Table \ref{ann_guideline} presents the distribution of SDoH for each split along with the IAA for all event types. The entirety of the test set and the majority of the validation set are doubly annotated. 

\subsection{SDoH Information Extraction}
We experimented with various LLM types and learning strategies, including i) fine-tuning BERT, ii) fine-tuning T5, and iii) in-context learning with GPT-4. The generative model experimentation with T5 and GPT-4 explored multiple prompting strategies, including i) single-step text2event (Event), and ii) two-step question answering (2sQA). Both prompting approaches were explored with T5 through fine-tuning and GPT-4 through in-context learning.

\textbf{Fine-tuning BERT (mSpERT):} Following prior work in the N2C2 SDoH challenge \citep{lybarger20232022}, we use our high-performing, multi-label variation of the Span-based Entity and Relation Transformer model (mSpERT) \citep{eberts2019SpERT,lybarger2022mSpERT}\footnote{\url{https://github.com/Lybarger/sdoh_extraction}}, as the BERT baseline. mSpERT is a span-based extractor that jointly extracts entities and relations. In the PedSHAC extraction task, mSpERT assigns multiple labels to a given span and assumes all predictions for a given span are associated with the same event. As all PedSHAC arguments share the same span as the trigger, mSpERT did not generate any relation predictions between spans. 

\textbf{Fine-tuning T5 with single-step text2event Prompting (T5-Event):} Recent work \citep{lu-etal-2021-text2event, ma-etal-2023-dice} demonstrates that entity and relation extraction tasks can be reformulated into text2event tasks using generative encoder-decoder models like T5 \citep{t5,chung2022scaling} and decoder-only models like GPT-4 \citep{openai2023gpt4}. We map each event annotation to a structured text representation  \citep{romanowski2023extracting,lu-etal-2021-text2event}\footnote{\url{https://github.com/romanows/SDOH-n2c2/}}. 
Figure \ref{SDOH_generative_models} illustrates our  T5-Event approach. Input sequences included the entire social history section and a model instruction. The target sequence was a sequence of SDoH events containing trigger type and text span, followed by the required and optional arguments. The trigger span was repeated with its argument to associate the arguments with the trigger span. Multiple events in the output were separated with `AND' for parsing. T5-Event extracts all PedSHAC SDOH events for a social history section in one step.

\textbf{Fine-tuning T5 with two-step QA Prompting (T5-2sQA):} we utilize a two-step pipeline approach to first extract triggers spans \citep{ma-etal-2023-dice} and then resolve subtype labels through multiple-choice questions \citep{ma2023multichoice}. Figure \ref{SDOH_generative_models} illustrates our two-step approach. In step 1, the model input is a prompt specifying the target event type and the social history section text, and the model's desired output is a list of trigger spans associated with the target event type.  In step 2, we apply multi-choice QA to resolve the argument subtype labels for each identified trigger and each argument type relevant to the event type. The input prompt specifies the argument, the relevant trigger within the note, and all possible argument subtypes. An additional choice, ``none,'' is added for optional arguments, indicating the argument may not be present for that event. The model output is the selected subtype.

\textbf{GPT-4 with In-context Learning:}  Previous research demonstrates LLMs can achieve high performance through in-context learning \citep{agrawal-etal-2022-large}. Additionally, some proprietary LLMs, including GPT-4, cannot currently be fine-tuned. Using prompt-based, in-context learning, information about the desired task is conveyed through instructions and few-shot examples. The larger context window of recent LLMs, including GPT-4 \citep{openai2023gpt4}, which can accommodate up to 32k tokens, allows detailed text-based instructions and several response examples to be included in the prompt.  We explored three in-context learning strategies: i) \textbf{Event} and \textbf{2sQA} -- simple instructions without explanation of annotated phenomena. For \textbf{GPT-Event}, our instruction contained a list of all the event and argument types and an illustration of the T5-Event output format using a randomly chosen example note. \textbf{GPT-2sQA} uses the same prompts provided to T5-2sQA, ii) \textbf{GPT + guide} -- \textit{2sQA} prompt with a brief description of target trigger/argument based on a summary of the annotation guideline, iii) \textbf{GPT + 3-shot} -- three few-shot examples, in addition to the \textit{GPT + guide} prompts.
For the \textit{+3-shot} setting, we randomly selected three example social history sections from the train set per GPT query, with some restrictions: (1) for trigger extraction: the three example notes contained zero, one, and more than one triggers of specific event type respectively;  (2) for required argument extraction, three randomly selected examples of events with that argument type (positive examples); and (3) for optional argument extraction such as \textit{residence}, one random negative example as an event without that argument, and two random positive examples, are included from event associated with the argument type.

\subsection{Experimental Paradigm}
In fine-tuning, we trained extraction models on the train set, optimized the hyperparameters on the validation set, and applied the best-performing models to the withheld test set. In in-context learning, we utilized the annotation guideline and examples from the train set. We initialized the BERT-based mSpERT model from Bio+ClinicalBERT \citep{alsentzer-etal-2019-publicly}. For T5 experimentation, we initialized from Flan-T5-Large (780M) \citep{chung2022scaling}, an instruction-tuned T5 variant.  For GPT-4 experiments, we used OpenAI's GPT-4-32k (version: 2023-03-15-preview) with the chat completion API provided through our HIPAA-compliant Azure server instance and utilized the `role' preset (`system', `user', and `assistant') arguments for providing our prompts. The system message includes the same instructions as the T5 experiments (except for the subtype options) and the distilled annotation guideline. The user message includes the note and subtype options for the argument extraction. We utilize multiple user-assistant input pairs to simulate the conversation history as in-context learning few-shot examples.

%% file: tables/annotation_guideline.tex
\resizebox{\textwidth}{!}{
\begin{tabular}{lllcccc}
\hline
\multirow{2}{*}{\textbf{Event}} & \multirow{2}{*}{\textbf{\begin{tabular}[c]{@{}l@{}}Trigger \\ \& Arg.\end{tabular}}} & \multirow{2}{*}{\textbf{\begin{tabular}[c]{@{}l@{}}Trigger examples\\ \& Argument subtypes\end{tabular}}} & \multicolumn{3}{c}{\textbf{\# labels}} & \multirow{2}{*}{\textbf{\begin{tabular}[c]{@{}c@{}}IAA\\ F1\end{tabular}}} \\ \cline{4-6}
 &  &  & \textbf{Train} & \textbf{Validation} & \textbf{Test} &  \\ \hline
Adoption & Trigger & ``adopted", ... & 27 & 4 & 9 & 100.0 \\ \hline

\multirow{2}{*}{\begin{tabular}[c]{@{}l@{}}Education \\ Access\end{tabular}} & Trigger & ``5th grade" , ``junior year", ... & 227 & 35 & 74 & 80.0 \\ \cline{2-3}
 & Status & (yes,no) & 227 & 35 & 74 & 80.0 \\ \hline
\multirow{2}{*}{Employment} & Trigger & ``Employment: ... ", ``works", ... & 390 & 45 & 117 & 81.1 \\ \cline{2-3}
 & Status & \begin{tabular}[c]{@{}l@{}}(employed, unemployed, retired, \\on disability, student, homemaker)\end{tabular} & 390 & 45 & 117 & 77.8 \\ 
 \hline %\cline{2-3} 
 
\multirow{2}{*}{Food Insecurity} & Trigger & ``food stamps", ``food insecurity", ... & 37 & 5 & 8 & 40.0 \\ \cline{2-3}
 & Status & (current, past, none) & 37 & 5 & 8 & 40.0 \\ \hline
 
\multirow{4}{*}{\begin{tabular}[c]{@{}l@{}}Living\\ Arrangement\end{tabular}} & Trigger & ``lives", ``foster care", ... & 676 & 101 & 195 & 90.4 \\ \cline{2-3}
 & Status & (current, past, future) & 676 & 101 & 195 & 88.5 \\ \cline{2-3}
 & Type* & \begin{tabular}[c]{@{}l@{}}(with both parents, with single\\ parent, with other relatives, with\\ foster family, with strangers)\end{tabular} & 566 & 86 & 160 & 88.4 \\ \cline{2-3}
 & Residence* & (home, institution, homeless) & 136 & 22 & 38 & 38.1 \\ \hline
\multirow{3}{*}{Mental Health} & Trigger & ``depression", ``self-harm", ... & 45 & 11 & 15 & 66.7 \\ \cline{2-3}
 & Status & (current, past, none) & 45 & 11 & 15 & 53.3 \\ \cline{2-3}
 & Experiencer & (patient, parent/caregiver) & 45 & 11 & 15 & 66.7 \\ \hline
\multirow{3}{*}{\begin{tabular}[c]{@{}l@{}}Substance Use \\ - Alcohol / \\ Drug / Tobacco \end{tabular}} & Trigger & ``meth", "alcohol", ``smokes",... & 265 & 38 & 78 & 86.4 \\ \cline{2-3}
 & Status & (current, past, none) & 265 & 38 & 78 & 85.7 \\ \cline{2-3}
 & Experiencer & (patient, parent/caregiver) & 265 & 38 & 78 & 73.2 \\ \hline
\multirow{3}{*}{Trauma} & Trigger & ``mentally abusive", ``bullying", ... & 132 & 23 & 33 & 88.9 \\ \cline{2-3}
 & Status & (yes, no) & 132 & 23 & 33 & 88.9 \\ \cline{2-3}
 & Type & \begin{tabular}[c]{@{}l@{}}(divorce / separation, loss, \\ psychological, physical, domestic\\ violence, sexual)\end{tabular} & 132 & 23 & 33 & 84.6 \\ \hline
\end{tabular} 
}

%% file: sections/5_results.tex
\subsection{Trigger and argument evaluation}

\subfile{../tables/performance_F1only}

\label{trigger_argument_results}
Following the evaluation criteria described in Section \ref{evaluation_criteria}, we report the extraction performance on the withheld PedSHAC test set in Table \ref{tab:performance} under two settings: i) fine-tuning with mSpERT and T5 and ii) in-context learning with GPT-4. We validate the F1 scores and assess significance using a pairwise non-parametric test (bootstrap test, p-val < 0.05) \cite{bergkirkpatrick2012} for all approaches, but only present a subset of significance testing results in Table \ref{tab:performance} due to lack of space. We consider the mSpERT model as a baseline for all approaches, with GPT-Event and GPT-2sQA base as a baseline for in-context learning approaches. The  `*' indicates performance fine-tuning approaches with significance over mSpERT  or vice
versa and  \textsuperscript{\textdagger} marks in-context learning models with significantly higher performance than GPT-Event and GPT-2sQA base. The highest performance in each row is boldfaced.

\textbf{Comparing performance against human IAA\footnote{Note that the last round IAA is not directly comparable to LLM performance. Because (1) IAA is from the last double-annotation round, while the model performance is calculated on 
the whole test set, (2) the test set has resolved the annotator disagreement from the IAA. Therefore, the IAA is not an upper bound for LLM performance on the test set, but a reference to `good' performance.}}, GPT+3-shot shows comparable performance in trigger micro average (82.3 F1) to corresponding IAA (85.1 F1), and T5-2sQA shows argument micro average (78.4 F1) close to corresponding IAA (80.0 F1).  For event types with lower IAA rates, such as \textit{Mental Health} (trigger and all arguments) and \textit{Living Arrangement} (\textit{residence} argument), the extraction performance is also lower, indicating complexity in the SDoH descriptions.

\textbf{For fine-tuning approaches}, all models exhibit high trigger extraction performance with no significant difference. 
%Given that arguments characterize the SDoH event types with useful subtypes, resolving them using the fine-tuning approaches could improve the integration of the extracted information into structured EHR data fields and further downstream clinical use. 
Comparing arguments micro average, T5-2sQA demonstrates significantly better performance than mSpERT, as well as all other in-context learning models. But on the level of individual argument types, T5-2sQA performance is similar to mSpERT and T5-Event, with the exception of the \textit{Living Arrangement} - \textit{type} argument. 
We observed no significant difference between T5-Event and T5-2sQA, indicating with sufficient fine-tuning data, the Flan-T5-large model can extract multiple events with complex, fine-grained event annotations appearing at the same time.

\textbf{Comparing in-context-learning approaches with GPT-4}, GPT-Event and GPT-2sQA base approaches demonstrate relatively lower performance when limited scheme information is incorporated into the prompt. Similar to the  T5-Event and T5-2sQA models, the GPT-Event and GPT-2sQA base approaches have no significant difference in the trigger and argument extraction performances.  Starting from GPT-2sQA base, adding the guidelines (+guide) provides the model with a detailed annotation scheme description, leading to significant improvement as 8.5 (from 71.3 to 79.8) among triggers and 9.8 (from 60.0 to 69.8)
among arguments. Adding three in-context learning examples further improves the performance (GPT+3-shot) from the base 2sQA with 11.0 (from 71.3 to 82.3) among triggers and 11.6 (from 60.0 to 71.6) among arguments. Adding the guidelines to the GPT-2sQA model (+guide) shows comparable trigger performance with the fine-tuned models. The GPT+3-shot achieves the highest trigger extraction performance, albeit without statistically significant improvement from the GPT+guide. Specifically, the GPT+3-shot model shows a significant increase in performance for \textit{Education access}, \textit{Employment}, and \textit{Substance Use} extraction over GPT-Event and GPT-2sQA base, while showing a significant increase even over mSpERT for \textit{Employment} extraction. The GPT+3-shot model demonstrates similar performance to the fine-tuned models for extracting \textit{Education Access}, \textit{Employment}, \textit{Living Arrangement}, and \textit{Substance Use} event types.

\subfile{../tables/performance_event}

\subsection{Event-level evaluation}
We additionally assess performance using a more rigorous \textit{event-level} evaluation criteria, which requires the equivalence (defined in Section \ref{evaluation_criteria}) of all arguments in an event type. A predicted event is considered correct if and only if its trigger overlaps with a trigger in the gold standard and all arguments in the event are correctly identified with the correct subtype labels. Table \ref{tab:eventF1} presents the event-level performance for the best GPT-2sQA approach and the rest of the approaches. We conduct the same pairwise significance testing across all models as Section \ref{trigger_argument_results}, yet exclude the results from Table \ref{tab:eventF1} to improve readability.

The T5-2sQA model achieves the highest micro-average performance, as well as significantly better performance than the in-context learning approaches in \textit{Living Arrangement}, \textit{Substance Use}, and micro average. Both mSpERT and T5-Event have similar performance to T5-2sQA. There is no significant difference among all fine-tuning models in any event. 

Note that the trigger extraction performance bounds event-level performance. Comparing Table \ref{tab:performance} with Table \ref{tab:eventF1}, three fine-tuning approaches have a relatively small performance drop on the micro average from trigger to event, as 6.2 (from 80.9 to 74.7) for T5-2sQA, 8.0 (from 79.6 to 71.6) for mSpERT and, 9.1 (from 79.5 to 70.4) for T5-Event. This is because trigger extraction is a more challenging task, and the fine-tuning-based LLMs can correctly predict the argument if they are able to correctly identify the trigger. This demonstrates great promise for fine-tuning-based LLMs' downstream clinical use at the event extraction level. On the other hand, the GPT+3-shot shows a performance drop of 28.3 (from 82.3 to 54.0). This is mainly because the GPT+3-shot model shows poor performance on some arguments (i.e. \textit{Living Arrange} - \textit{residence}) and the difficulty of predicting multiple arguments correctly at the same time for the same event.

% oh no, this sentence is not about the trigger, it is about the argument, this should go to the overall error analysis.
%This is because it sometimes extracts meaningful SDoH information but fails to overlap with the gold annotation, especially in the \textit{Food insecurity} events. For example, clinicians tend to follow a template format: `Food insecurity: NO'. while GPT+3-shot tends to extract the phrase following the prefix and predicts 'No' as the trigger, the annotators annotate the prefix, 'Food insecurity', as the gold trigger. 

\subsection{Error Analyses}
Comparing errors across different learning strategies, we observed that the fine-tuning models tend to have relatively lower recall than precision, while the in-context learning models tend to have lower precision than recall. While fine-tuning models perform well in extracting SDoH for event types well-represented in the training set, they demonstrate relatively poorer generalizability. This could be because fine-tuning models contain much fewer parameters than GPT-4 and have less prior knowledge about some SDoH factors. For example, if a \textit{Mental Health} trigger phrase is uncommon and not previously seen in the train set, the fine-tuning models can fail to extract it. On the other hand, the in-context learning approaches tend to interpret SDoH extraction in a broader context and extract events outside the annotation scheme. For example, `Dad </name>, Mom </name> and Sister </name>' is a list of the family members' names, which does not explicitly state the patient's living arrangement. However, the GPT+3-shot approach considers this span implying a \textit{Living Arrangement} event and annotates it as a trigger.
%with the argument \textit{residence} subtype as \textit{home}.

Without fine-tuning, GPT+3-shot is very sensitive to the instructions provided in the form of the guideline. For example, our guideline did not state that the \textit{residence} subtype needs to be explicitly mentioned, and GPT-4 predicted descriptions such as `lives with parents' having the optional argument \textit{residence} with the subtype \textit{home}'. Such false positives resulted in a precision of 17.2 and 28.6 F1 for the \textit{residence} argument. GPT+3-shot also sometimes extracts meaningful SDoH information but fails to overlap with the gold annotation, especially in the \textit{Food Insecurity} events. For example, clinicians tend to follow a template format: `Food insecurity: NO'. while GPT+3-shot tends to extract the phrase following the prefix and predicts 'No' as the trigger, the annotators annotate the prefix, `Food insecurity', as the gold trigger. On the other hand, because T5-based approaches learn % a decision function 
from abundant annotated data, they were able to learn from the actual implementation of the guide and implicitly understand edge cases that are not explicitly defined in the guide. Future GPT-based models could use better-designed prompts to incorporate more detailed instructions or better sample selection approaches for in-context learning.

Consistent with errors identified by prior work \citep{ji2023survey}, both generative models (T5 and GPT-4) show a problem of hallucination \citep{ji2023survey}, outputting with improper formats, which range from minor modifications to spacing, punctuation, and casing. Another type of hallucinated response is spans that do not correspond to the original text, such as synonyms to the original SDoH determinants. %Additionally, these generative models sometimes generate synonyms to the actual spans. 
%We consider predictions that fail to map to the gold social history section and the annotation scheme as invalid. 
 We consider the generated output invalid if the predictions do not comply with the predefined output format or the predictions contain predicted spans that do not exactly match the original text.
We observed a 3-5\% invalid rate for trigger prediction and less than 1\% for argument prediction in the QA approaches. Future work could apply approaches to better constrain the prediction within the note and annotation scheme, including rule-based post-editing such as minimum edit distance, self-verification \citep{gero2023self} and constrained decoding \citep{lu-etal-2021-text2event}.

\subsection{Limitations}

Our annotation of the SDoH events in PedSHAC is limited to a single hospital system and its pediatric population. The distribution of the SDoH events may not be representative of other pediatric populations. The relatively lower frequencies of some of the event types may result from the patient population at our institution. The current annotation scheme does not allow multiple events of the same event type to have the same trigger span. For example, in the sentence, `He lives with grandma first, and then with his parents', both \textit{past} and \textit{current} \textit{Living Arrangement} events should have the same trigger 'lives' but is not allowed. In future work, we plan to modify the annotation scheme to allow multiple events of the same type associated with the same trigger. Some downstream clinical research may need even more fine-grained annotation.

%% file: tables/performance_F1only.tex
\begin{table*}[ht] %ht
\centering
\small
\setlength{\tabcolsep}{6pt}
\begin{tabular}{lrcccc|cccc}
\hline
\multirow{4}{*}{\textbf{Event}} & \multirow{4}{*}{\textbf{\begin{tabular}[c]{@{}c@{}}Trigger\\ \& Arg.\end{tabular}}} & \multirow{4}{*}{\textbf{\begin{tabular}[c]{@{}c@{}}\# gold \\ labels\end{tabular}}} & \multicolumn{7}{c}{\textbf{Extraction performance   (F1)}} \\ \cline{4-10} 
 &  &  & \multicolumn{3}{c}{\textbf{Fine-tuning}} & \multicolumn{4}{c}{\textbf{In-context learning}} \\ \cline{4-10} 
 &  &  & \multirow{2}{*}{\textbf{mSpERT}} & \multirow{2}{*}{\textbf{\begin{tabular}[c]{@{}c@{}}T5-\\      Event\end{tabular}}} & \multirow{2}{*}{\textbf{\begin{tabular}[c]{@{}c@{}}T5-\\      2sQA\end{tabular}}} & \multirow{2}{*}{\textbf{\begin{tabular}[c]{@{}c@{}}GPT-\\      Event\end{tabular}}} & \multicolumn{3}{c}{\textbf{GPT-2sQA}} \\ \cline{8-10} 
 &  &  &  &  &  &  & \textbf{base} & \textbf{+guide} & \textbf{\begin{tabular}[c]{@{}c@{}}+guide\\      +3-shot\end{tabular}} \\ \hline
Adoption & Trigger & 9 & \textbf{84.2} & 82.4 & \textbf{84.2} & 58.1 & 66.7 & 66.7 & 54.5 \\
 \hline
\multirow{2}{*}{Edu. Access} & Trigger & 74 & 78.0 & 79.1 & 84.1 & 71.6 & 75.9 & 84.9& \textbf{85.7\textsuperscript{\textdagger}} \\
 & Status & 74 & 78.0 & 79.1 & 84.1 & 71.6 & 53.3 & \textbf{85.5\textsuperscript{\textdagger}} & 84.5\textsuperscript{\textdagger} \\
  \hline
\multirow{2}{*}{Employment} & Trigger & 117 & 75.1 & 78.9 & 81.1 & 69.1 & 73.4 & 85.5*\textsuperscript{\textdagger} & \textbf{89.2*\textsuperscript{\textdagger}} \\
 & Status & 117 & 71.4 & 76.3 & 74.3 & 60.8 & 64.0 & 76.9\textsuperscript{\textdagger} & \textbf{80.6*\textsuperscript{\textdagger}} \\
 \hline
\multirow{2}{*}{Food Insecurity} & Trigger & 8 & \textbf{93.3} & 87.5 & \textbf{93.3} & 53.3 & 0.0 & 70.0 & 87.5 \\
 & Status & 8 & \textbf{93.3} & 87.5 & \textbf{93.3} & 53.3 & 0.0 & 70.0 & 87.5 \\
  \hline
\multirow{4}{*}{Living Arrg.} & Trigger & 195 & 84.8 & \textbf{86.5} & 85.4 & 82.3 & 80.9 & 83.7 & 84.0 \\
 & Status & 195 & 82.6 & 83.4 & \textbf{84.4} & 80.2 & 78.4 & 81.0 & 78.4 \\
 & Type & 160 & 83.3 & 82.7 & \textbf{88.7*} & 76.6 & 75.4 & 81.2 & 77.9 \\
 & Residence & 38 & 63.5 & \textbf{67.6} & 62.2 & 27.7 & 27.2 & 28.0 & 28.6 \\
  \hline
\multirow{3}{*}{Mental Health} & Trigger & 15 & 38.1 & 25.0 & 36.4 & 26.3 & 51.9 & \textbf{53.3} & 51.6 \\
 & Status & 15 & 28.6 & 25.0 & 34.8 & 26.3 & 35.7 & 38.7 & \textbf{43.8} \\
 & Experiencer & 15 & 9.5 & 8.3 & 17.4 & 21.1 & 35.7* & 40.0* & \textbf{43.8*} \\
  \hline
\multirow{3}{*}{Subst. Use} & Trigger & 78 & \textbf{85.5*} & 81.6 & 81.9 & 54.1 & 64.2 & 73.5\textsuperscript{\textdagger} & 80.2\textsuperscript{\textdagger} \\
 & Status & 78 & 81.4 & 78.1 & \textbf{81.9} & 50.8 & 63.2 & 69.0 & 76.8\textsuperscript{\textdagger} \\
 & Experiencer & 78 & 74.5 & 80.3 & \textbf{80.6} & 49.2 & 63.2 & 72.1\textsuperscript{\textdagger} & 80.0\textsuperscript{\textdagger} \\
 \hline
\multirow{3}{*}{Trauma} & Trigger & 33 & 62.1 & 54.5 & 53.3 & 58.6 & 5.7 & 55.3 & \textbf{70.2} \\
 & Status & 33 & 51.7 & 54.5 & 54.2 & 58.6 & 5.7 & 55.3 & \textbf{63.2} \\
 & Type & 33 & 55.2 & 51.5 & 54.2 & 55.2 & 5.7 & 55.3 & \textbf{66.7} \\ \hline
\multirow{2}{*}{Micro Avg} & Trigger & 529 & 79.6 & 79.5 & 80.9 & 69.9 & 71.3 & 79.8\textsuperscript{\textdagger} & \textbf{82.3}\textsuperscript{\textdagger} \\
 & Arguments & 844 & 75.3 & 76.0 & \textbf{78.4*} & 62.0 & 60.0 & 69.8\textsuperscript{\textdagger} & 71.6\textsuperscript{\textdagger} \\ \hline
\end{tabular}
\caption{Model performance F1 (\%) on event triggers and arguments from the PedSHAC withheld test set. The asterisk * indicates that performance was significantly better (p<0.05) than mSpERT or vice versa. The symbol \textsuperscript{\textdagger} marks in-context learning models with significantly higher performance than GPT-Event and GPT-2sQA. The highest performance in each row is in boldface.
}
\label{tab:performance}

\end{table*}

%% file: tables/performance_event.tex
% \usepackage{multirow}

\begin{table}[]
\small
\setlength{\tabcolsep}{2pt}
\resizebox{0.5\textwidth}{!}{% use resizebox with textwidth
\begin{tabular}{lcccccc}
\hline
\multirow{3}{*}{\textbf{Event}} & \multirow{3}{*}{\textbf{\begin{tabular}[c]{@{}c@{}}\\ \# \\ gold\\ labels\end{tabular}}} & \multicolumn{5}{c}{\textbf{Event extraction performance (F1)}} \\ \cline{3-7} 
 &  & \multicolumn{3}{c}{\textbf{Fine-tuning}} & \multicolumn{2}{c}{\textbf{\begin{tabular}[c]{@{}c@{}}In-context\\ learning\end{tabular}}} \\ \cline{3-7} 
 &  & \textbf{mSpERT} & \textbf{\begin{tabular}[c]{@{}c@{}}T5-\\  Event\end{tabular}} & \textbf{\begin{tabular}[c]{@{}c@{}}T5-\\  2sQA\end{tabular}} & \textbf{\begin{tabular}[c]{@{}c@{}}GPT-\\  Event\end{tabular}} & \textbf{\begin{tabular}[c]{@{}c@{}}GPT+\\ 3-shot\end{tabular}} \\ \hline % \cline{1-7}
 
Adoption & 9 & \textbf{84.2} & 82.4 & \textbf{84.2} & 58.1 & 54.5 \\
Edu. Acc. & 74 & 78.0 & 79.1 & 84.1 & 71.6 & \textbf{84.5} \\
Employment & 117 & 71.4 & 73.5 & 74.3 & 60.8 & \textbf{79.7} \\
Food. Insec. & 8 & \textbf{93.3} & 87.5 & \textbf{93.3} & 53.3 & 73.7 \\
Living Arrg. & 195 & 72.8 & 69.7 & \textbf{74.9} & 19.8 & 12.6 \\
Mental Health & 15 & 9.5 & 8.3 & 17.4 & 21.1 & \textbf{37.5} \\
Subst. Use & 78 & 75.9 & 75.0 & \textbf{80.6} & 45.9 & 78.0 \\
Trauma & 33 & 51.7 & 51.5 & 53.3 & 55.2 & \textbf{59.6} \\ \hline
\begin{tabular}[c]{@{}l@{}}Micro\\ Avg\end{tabular} & 529 & 71.6 & 70.4 & \textbf{74.7} & 42.6 & 54.0 \\ \hline
\end{tabular}
%\textsuperscript{\textdagger}

}
\caption{Model performance F1 (\%) with the \textit{event-level} evaluation on the PedSHAC withheld test set.} %The symbol \textsuperscript{\textdagger} marks the GPT+3-shot approach with significantly higher performance than GPT-Event. The highest performance in each row is in boldface.}
\label{tab:eventF1}
\end{table}

%% file: sections/6_conclusion.tex
In this work, we present a novel corpus, PedSHAC,  annotated for SDoH. Our corpus has 1,260 social history sections of pediatric patients annotated across 10 SDoH event types. %By summarizing and defining all those SDoH types, we make the measurement of SDOH in pediatric health more comprehensive.
We envision such fine-grained annotation on multiple critical SDoH types can help the research community study the impact of SDOH on other child health outcomes.
We explored LLM-based IE across multiple dimensions, including pre-trained architectures -- mSpERT, Flan-T5, and GPT-4; learning strategies -- fine-tuning and in-context methods; and prompting approaches -- one-step text-to-event and two-step QA.
%PedSHAC has an overall annotator agreement of 81.9 F1. Our highest-performing approach reaches 82.3 F1 in trigger extraction through GPT-4 in-context learning, and an 78.4 F1 in argument extraction through Flan-T5 fine-tuning. 
Our results demonstrate that detailed SDoH representations can be extracted from pediatric narratives with performance comparable to human annotators, providing an automatic approach for incorporating valuable SDoH information in clinical and research applications. 

Future work for the corpus development could include addressing the current limitations, through actual user studies to pinpoint the needs and possibly expanding the current SDoH annotation to encompass more hospital systems and pediatric subpopulations. We also plan to explore other IE approaches such as (1) using effective data selection strategies such as active learning \cite{lybarger2021annotating} in the annotation phase could help save annotation costs, (2) GPT-4 prompt-tuning including the involvement of medical experts, automatic prompt generation \citep{zhou2022large}, and self-verification \citep{weng2022large} to improve the response quality. 

Our proposed automatic IE approaches allow extracted SDoH information to be directly incorporated in EHRs in a tabular form, we envision our work to help downstream clinical applications through better quantifying the presence of various SDoHs in pediatric populations.

%% file: sections/7_ethics.tex
% ref: https://aclrollingreview.org/ethicsreviewertutorial

We obtained the necessary approvals from our institution's Institutional Review Board (IRB), with a waiver of patient content on using their clinical notes. SDoH sections in clinical notes may contain Protected Health Information (PHI) including potentially identifiable information, like names, occupations, contact information, and other identifiers. All researchers and annotators received the necessary human subjects training to interact with patient data, including PHI. We used secured servers to ensure data security. Our GPT-4 experiments were conducted on a Health Insurance Portability and Accountability Act  (HIPAA)-compliant Azure environment and ensured that no queries would be recorded by OpenAI. The patient populations in our corpus may not the representative of populations at other institutions or the broader population, which may inadvertently bias our extraction models and impact the generalizability. We believe our work will benefit the practice of automatic SDoH IE from pediatric narratives, as well as the general domain IE through LLMs.

%Check Casing abbreviations and links for references